# Augmenting The Weather:
# A Hybrid Counterfactual–SMOTE Algorithm for Improving Crop Growth Prediction When Climate Changes

Mohammed Temraz[1] & Mark T. Keane[2,3,4]

College of Computer & Cyber Sciences, University of Prince Mugrin, Madinah, Saudi Arabia

School of Computer Science, University College Dublin, Belfield, Dublin 4, Ireland

Insight Centre for Data Analytics, University College Dublin, Belfield, Dublin 4, Ireland

VistaMilk SFI Research Centre, University College Dublin, Belfield, Dublin 4, Ireland

mohammed.temraz@upm.edu.sa; 0000-0002-4542-6596

mark.keane@ucd.ie; 0000-0001-7630-9598

**Abstract**

In recent years, humanity has begun to experience the catastrophic effects of climate change as economic sectors (such as agriculture) struggle with unpredictable and extreme weather events. Artificial Intelligence (AI) should help us handle these climate challenges but its most promising solutions are not good at dealing with climate-disrupted data; specifically, machine learning methods that work from historical data-distributions, are not good at handling out-of-distribution, outlier events. In this paper, we propose a novel data augmentation method, that treats the predictive problems around climate change as being, in part, due to class-imbalance issues; that is, prediction from historical datasets is difficult because, by definition, they lack sufficient minority-class instances of "climate outlier events". This novel data augmentation method -- called Counterfactual-Based SMOTE (CFA-SMOTE) -- combines an instance-based counterfactual method from Explainable AI (XAI) with the well-known class-imbalance method, SMOTE. CFA-SMOTE creates synthetic data-points representing outlier, climate-events that augment the dataset to improve predictive performance. We report comparative experiments using this CFA-SMOTE method, comparing it to benchmark counterfactual and class-imbalance methods under different conditions (i.e., class-imbalance ratios). The focal climate-change domain used relies on predicting grass growth on Irish dairy farms, during Europe-wide drought and forage crisis of 2018.

# 1 Introduction

In recent years, many of us have personally experienced the effects of climate change in our daily lives whether it be a delayed flight caused by the smoke from forest-fires, a winter storm that wrecks our house or flooded roads caused by "rain bombs". Artificial Intelligence (AI) has been promoted as a key technology to combat climate change and to enhance sustainability, by improving our understanding and prediction of climate-disruptive events. For example, the United Nations "AI for Good" platform has promoted a variety of AI technologies under its Sustainability Goals [31]. In this paper, we report on AI attempts to predict crop growth in the face of climate change. Climate change presents fundamental challenges for current much-heralded, state-of-the-art AI methods, such as deep learners [16, 30], because it involves out-of-distribution events relative to historical data, producing distributional shifts that surface known weaknesses in these models.

In recent years, many new neural network techniques, so-called deep learning methods, have been advanced that show significant promise in solving hitherto unsolved problems, such as human-level image and video analysis (e.g., for medical diagnosis), the real-time control of machinery (e.g., self-driving cars) and communication using natural language (e.g., translation, question answering) [3, 7, 24, 26, 29 ]. However, though these models are impressive, their success fundamentally depends on the data they use; if that data is noisy, or if it is unrepresentative or somehow distributionally-incomplete, then these methods tend to fail. Bengio [2] has argued that many of these advanced models fail because they do not generalize well to new, unseen data or are not good at handling distributional shifts in target phenomena. He argues that deep learners rely on the "independent and identically-distributed assumption" (i.e., the i.i.d assumption); namely, the assumption that all new data encountered by the model is, essentially, drawn from the same distribution as the original training data. So, when this assumption is violated, the performance of these models suffers.

Unfortunately, climate change is an "i.i.d. violating" phenomenon; as climate change occurs, historical training data becomes progressively less representative of current events and, therefore, prediction becomes less accurate. Climate change introduces out-of-distribution data-points leading to distributional shifts relative to historical data; so, historical datasets that capture "normal" climate events (e.g., moderate, temperate rainfall in a given location) are supplanted by "abnormal" outlier events (e.g., extended periods of drought in the same location) requiring these distributional shifts to be handled. So, as well as disrupting the weather, climate change specifically disrupts the datasets required by state-of-the-art AI methods, uncovering known weaknesses in their operation. In this paper, we address the problem of using AI to predict crop growth during periods of climate disruption, specifically grass growth in dairy agriculture. We propose that the distributional issues caused by climate change, essentially, cause a class-imbalance problem for these predictive models (i.e., there are many normal-weather data-points, but few outlier-weather data-points), that can be solved by data augmentation techniques.

Therefore, in this paper, we propose a novel counterfactual data-augmentation method based on SMOTE algorithm, called *Counterfactual-Based SMOTE* (CFA-SMOTE) that creates synthetic data-points representing outlier, climate-events that are introduced to the dataset with a view to improving predictive performance. Specifically, the new hybrid method combines a popular data augmentation method from the class imbalance literature, called SMOTE (see [4]) and an instance-based counterfactual method, called Counterfactual Augmentation (CFA) (see [37]) to address the class imbalance problems caused by climate change. This instance-based counterfactual (CFA) method for oversampling is used to introduce a set of synthetic counterfactual instances in the minority class as an input for SMOTE, to improve the density of borderline area, with a view to improving prediction accuracy. Our hypothesis is that the use of CFA with SMOTE decreases the risk of generating noisy instances in the minority class. Technical details of the SMOTE algorithm, its variants, and other instance-based and optimization counterfactual techniques are introduced in the following sections. As such, in this paper, we report three key, novel contributions showing that:

- Prediction in the presence of climate change can be cast as a class-imbalance problem that is solvable using data augmentation methods.
- The combination of a counterfactual method with SMOTE (in CFA-SMOTE) leads to a better solution to these augmentation problems than that offered by benchmark counterfactual and SMOTE-based methods on their own.
- CFA-SMOTE works best, relative to these other methods, under the conditions of extreme imbalance ratios (which perhaps best reflect the current state of climate datasets).

In the remainder of this paper, we introduce the background to the current research and report on a series of experiments that show how this counterfactual data-augmentation method can improve model predictions. In the next section (Section 2), we describe why grass growth prediction is important to the dairy sector, a problem domain that has been impacted by climate change. Then, we describe the background AI research on computing counterfactuals and some popular approaches to the class imbalance problem. In Section 3, we detail the present data augmentation method, CFA-SMOTE. Finally, we present a series of novel experiments on using counterfactual data augmentation to improve grass growth prediction during periods of climate disruption (see Section 4).

## 2 Related Work

Related work in this topic area draws on three different strands of research: on (i) the problem domain of predicting grass growth on dairy farms, (ii) data-level sampling techniques for the class imbalance problem and (ii) counterfactual methods for data augmentation. In this section, each of these three topic areas are reviewed to relate them to the current work.

### 2.1 Growing Grass on Irish Dairy Farms

The present work reflects a collaboration, since 2016, between data scientists at University College Dublin (Ireland) and agricultural scientists at Teagasc (Ireland's Agriculture and Food Development Authority) exploring the development of Smart Agriculture technologies in several key problem domains (e.g., food analysis, animal disease diagnosis, crop growth prediction). The present project focused on crop group prediction using AI techniques to predict grass growth on Irish dairy farms, a $5.4b agricultural sector. In Ireland, dairy farming is mainly pasture-based; animals are grass-fed outdoors for most of the year, rather than being housed indoors and fed on imported supplements (e.g., soya, cornmeal). The success of this more sustainable type of farming depends crucially on proper pasture management; farmers need to budget their grass, as they rotate their herds over different paddocks (the term used for "fields"), resting some fields, grazing others and spreading fertilizer when appropriate. Teagasc provides an online, decision-support tool for dairy farmers to do grass budgets, called PastureBase Ireland (PBI; www.pbi.ie) [11]. The PBI tool models a farm's paddocks, its cattle herd and expected grass growth to project future forage levels. Predicted grass growth is a key variable in this budgeting system and, traditionally, is estimated using mechanical models (see MoSt in [32]). An alternative approach has recently been explored

using AI methods on the historical data entered by farmers using the PBI system (e.g., involving 100,000+ data-entries over 6 years); each data-point represents a specific farm on a given day, recording the features *farm-id, time-of-year*, *current-grass-cover* (i.e., amount of dry biomass on the farm above 4cm grass height) and 3 weather parameters (i.e., *rainfall, temperature,* and *solar-radiation*). Several different AI models, using Bayesian techniques, *k*-NNs and neural networks, have shown good predictive performance using this data, but accuracy tends to suffer during periods of climate disruption (see [21, 22, 37]). For example, in 2018, most of Europe experienced a very hot summer, with drought conditions causing forest fires in Scandinavia and a forage crisis in Ireland (see Fig. 1 for 2018's temperature deviations). Although grass growth generally increases as temperatures increase, high temperatures and solar radiation can combine with low soil-moisture to burn grass, effectively leaving farmers with no feed for their cattle.

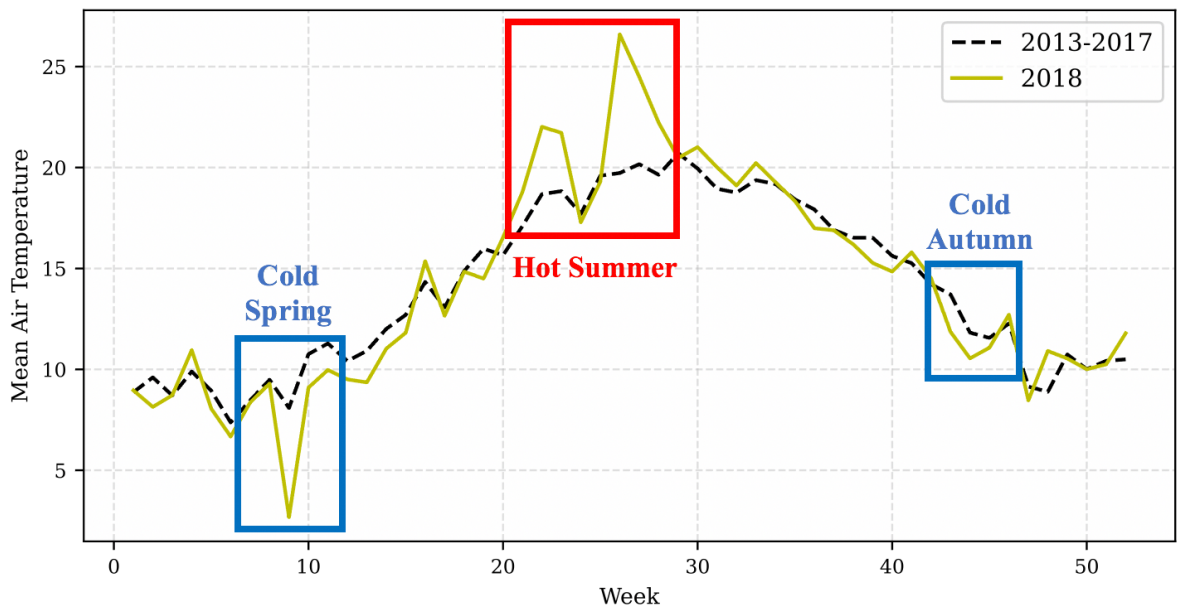

**Fig. 1** Air Temperature in Ireland (2013-2018): In 2018, mean air temperatures deviated significantly from long-term mean temperatures (2013-2017), with a cold spring (in March), a hot summer (in July) and a cold autumn (in October)

Analyses of the basis for the 2018-predictions of a normally-accurate *k*-NN model showed that accuracy depended on the presence of extreme-weather outliers in the training data; that is, the model relied on *past extreme-weather instances* to predict growth when *climate-disruptive test-cases* were encountered [37]. However, by definition, these extreme-weather cases were rare in the historical data; they were scarce outlier events dotted amongst normal weather patterns (e.g., see Figs. 2 and 3). So, the failure of these models was shown to be, in part, due to the lack of enough historical data about extreme-weather events [37]. Temraz et al. [37] showed that if extreme-weather cases were augmented using counterfactual XAI methods then predictions improved relative to a dataset using known extreme-weather cases; but, they did not test the overall predictive performance of the system and neither did they compare performance to that found for traditional data augmentation methods (such as SMOTE).

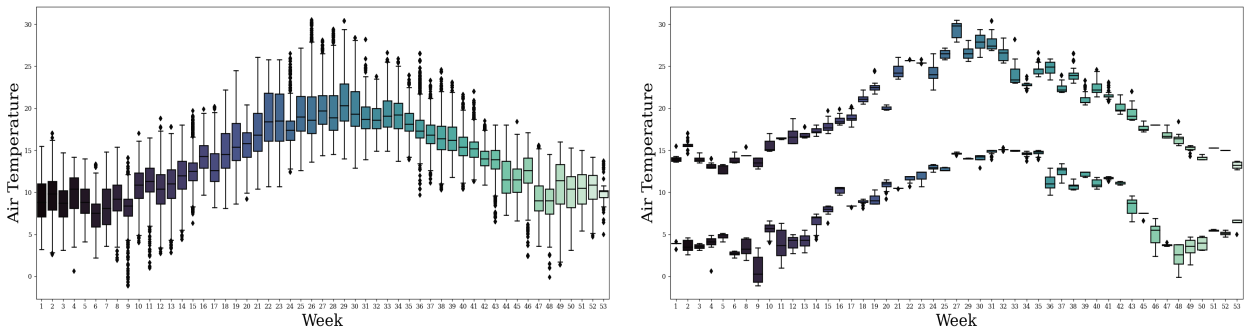

**Fig. 2** Distribution of temperature values, with box plots, showing weekly means for 2013-2018, for all the data (left) and for the high/low climate-disruptive outliers extracted from that data (right)

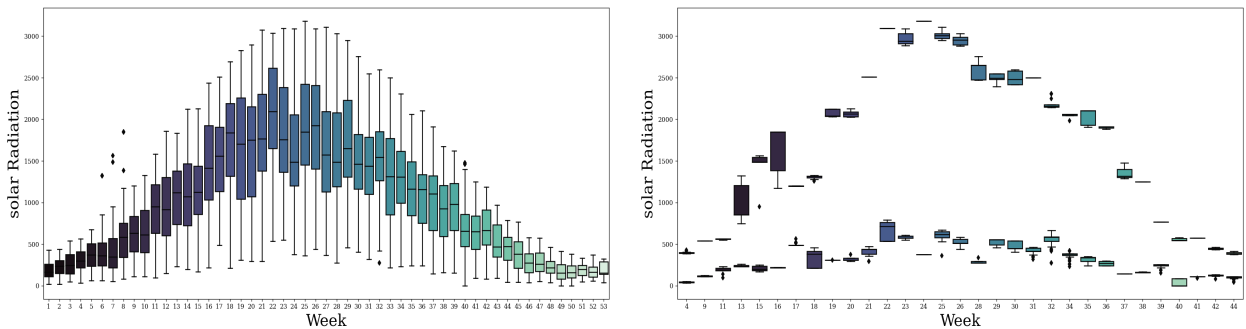

**Fig. 3** Distribution of solar-radiation values, with box plots, in weekly means for 2013-2018, for all the data (left) and for high/low climate-disruptive outliers extracted from that data (right)

Technically, these results suggest that these models fail to handle climate change because of a class imbalance problem; that is, the historical data divides into a minority class of extreme-weather instances that is underpopulated relative to the majority class of normal-weather instances in a way that undermines performance. The hypothesis being that, if the minority extreme-weather class could be populated with more data, perhaps generated synthetically, then the model's performance on future climate-disruptive events should improve. As no previous work has specifically tested this hypothesis, we test it here using the *Counterfactual-Based SMOTE* (CFA-SMOTE) algorithm. But, before reporting these tests, we first review the background to class imbalance problems, the sampling methods used to solve it and the novel counterfactual method proposed here.

## 2.2 Class Imbalance: Problem & Methods

Many machine learning models show poor performance when datasets have an imbalance between an under-populated minority class (e.g., of rare instances) and a majority class (e.g., of common instances). Class imbalance problems occur in many domains [23, 36, 41]; for instance, in fraud detection (where fraudulent transactions are rarer than valid ones) and medical diagnosis (rare illnesses that are missed). A key novelty of the current work is the design decision to cast the prediction difficulties caused by climate change as a being due to a class imbalance problem between a *minority class of extreme-weather instances* and a *majority class of normal-weather instances* (defined using a $2\sigma$ statistical boundary to divide the classes). To re-balance a dataset, class-imbalance methods typically down-samples the majority class (e.g., Random Under-Sampling [15]) or up-samples the minority class (e.g., Random Over-Sampling [25]). In Random Under-Sampling, the class distribution is balanced by randomly selecting and removing instances from the majority class. However, one drawback of this method is that, as it randomly removes instances from the majority class, data can be discarded that may be useful. On the other hand, Random Over-Sampling seeks to balance the class distribution by randomly adding certain copies of some of the minority classes to the training data. In a similar vein, although this method can balance the dataset, it has some drawbacks. As it is merely using the same instances in the minority class, no new information is added to the dataset and, hence, it can lead to overfitting. The third option – SMOTE – embraces a somewhat different approach by oversampling from the minority class.

Synthetic Minority Oversampling TEchnique or SMOTE [4, 9] is a very popular and successful solution to class-imbalance problems. SMOTE oversamples the minority class by interpolating between nearest neighbours in that class, to create "synthetic" instances. By interpolating instead of copying instances, SMOTE reduces the risk of overfitting. Briefly, SMOTE works as follows. Assume that $P$ is the positive (minority) class and $N$ is the negative (majority) class in the dataset $T$ (where $T = P \cup N$). In its first step, SMOTE randomly selects a minority instance, $p_i$, from the positive (minority) class, $P$, and then determines $m$, as the nearest neighbors of $p_i$. After determining $m$ nearest neighbors of $p_i$, SMOTE selects a random neighbor $m'$ (where $m' \in P$). Finally, a new instance, $p_{new}$, is generated in the minority class using the following formula: $p_{new} = p_i + (m' - p_i) \times \delta$ (where $\delta \in [0,1]$). However, one of the potential issues with SMOTE is that it generates synthetic minority instances without reference to the majority class. Furthermore, in its original form it did not consider the possibility that some instances in the minority class could be better than others to use in the data generation process. Consequently, several improvements to SMOTE have been proposed that priorities some minority instances over others.

Notably, Borderline-SMOTE [10] creates synthetic instances by only using minority instances that are close to the decision boundary between the classes. In B-SMOTE, for a given query instance in the minority class, if the set of its majority nearest neighbors is larger than that of its minority ones, this query instance will be easily misclassified and put into a DANGER set. This DANGER set holds the borderline instances of the minority class for the query instance, and the steps from the standard SMOTE method are applied to them to generate synthetic instances in the minority class. B-SMOTE improves on the performance seen for SMOTE on many datasets. Recently, Temraz et al. [37] have shown that counterfactual methods from eXplainable AI (XAI) can also be used to supplement instances close to the decision boundary in a way that might also improve SMOTE; though they did not test this proposal. Like B-SMOTE, this counterfactual method focuses on instances close to the decision boundary, though it differs in generating synthetic instances from historical counterfactuals in the dataset.

Some methods explore different ways to define regions within which to generate minority instances (e.g., Geometric-SMOTE and ADASYN). Douzas & Bacao [8] proposed Geometric-SMOTE, that defines a geometric region (named hyper-spheroid) by identifying a safe area around each instance in the minority class for generating synthetic instances. He et al. [14] proposed Adaptive Synthetic (ADASYN), that generates instances in the minority class according to their distributions. Specifically, the method generates new synthetic instances from minority instances that are harder to learn compared to minority instances that are easier to learn (where ease-of-learning represents the total number of instances in the *k*-nearest neighbors that belong to the majority class). All of these methods work by creating synthetic instances to add to the minority class, to re-balance the classes and improve prediction accuracy.

## 2.3 Counterfactuals: From XAI To Data Augmentation

In recent years, there has been many research initiatives to alleviate concerns of transparency and trust in AI by making AI systems more interpretable and explainable to humans by providing explanations; so-called eXplainable Artificial Intelligence (XAI) [17, 20, 38]. Counterfactual algorithms have recently been heavily researched in XAI [5, 6, 17, 19, 20, 27, 33, 34, 38], where they provide explanatory contrasts to an original query item; for instance, if you receive an automated refusal of a loan of $10k and ask "why?", a counterfactual explanation might reply "Well, if you asked for a lower loan of $9k, you would have been granted the loan". Counterfactuals tell us about the feature-value changes that lead to different outcomes (e.g., $10k to $9k), flipping a predicted outcome from one class to another.

Wachter et al.'s [38] seminal paper on XAI proposed a perturbation-based method to counterfactual generation optimization techniques, borrowed from adversarial learning. These techniques find counterfactual-instances to a test-instance in a generated space of perturbations of the test-instance, under a loss function using a scaled $L1$-norm distance-metric, where the proximity of the counterfactual-instance to the test-instance is balanced against its proximity to the decision boundary for the counterfactual class. Mothilal et al. [27] improved on this perturbation method, with some additional constraints on sparsity and diversity in a method called *Diverse Counterfactual Explanations* (DiCE). DiCE perturbs feature-values and then filters its results based on broad constraints of proximity and diversity. Although this method can generate diverse counterfactuals, it has some drawbacks. Given its ''blind'' perturbation of test-items, like all perturbation methods it can sometimes generate out-of-distribution, invalid data-points. In the present tests, we use DiCE as a comparator to our counterfactual method, as it has become a benchmark-method for tests of counterfactual generation.

Although there are 150+ counterfactual algorithms in the XAI literature, very few of these papers consider using these methods for data augmentation (see [12, 13, 18, 28, 35, 37, 40]). Most of these methods have used counterfactual methods to augment datasets to determine if performance improvements can be found in Reinforcement Learning [28], in adjusting textual data with counterfactuals [18, 40] and in dealing with dataset shifts [35]; however, they tend to user their own bespoke-methods rather than the well-tested, benchmark methods from XAI. However, Hasan & Talbert [13] and Temraz et al. [37] do use key XAI methods. Hasan & Talbert [13] used DiCE to test whether augmentation using counterfactual XAI could improve predictive performance, and found minimal improvements. However, they did not compare DiCE's performance to other counterfactual methods or, indeed, other class-imbalance methods (such as SMOTE). Temraz et al. [37] also used DiCE and compared it to an instance-based method they called Counterfactual Augmentation (following Keane & Smyth [19]). However, they did not compare performance to other class-imbalance methods, and they did not test *overall* predictive performance. Temraz et al. [37] compared the performance of known, minority-class-instances against counterfactually-generated minority-class-instances, which was a very circumscribed test of these methods in the climate domain; that is, the rest of the dataset was not used in these tests and all indications suggested that overall performance improved minimally (as Hasan [12] found). Furthermore, again, these papers did not consider counterfactual augmentation relative to the rich, existing literature on class-imbalance methods, such as the many SMOTE-variants proposed.

In this paper, we advance a novel method, called *Counterfactual-Based SMOTE* (CFA-SMOTE) that combines an instance-based counterfactual method (from Temraz et al., [37]) with the SMOTE algorithm. As we have seen, prior work in this climate domain suggests that counterfactual augmentation methods on their own do not deliver significant performance improvements. Hence, our proposal of a method that attempts to mix the best of counterfactual and class-imbalance methods with a view to determining how well a combined method might work.

## 3  The Counterfactual-Based SMOTE (CFA-SMOTE)

As we have seen, the major novel focus of the current paper is on whether counterfactual methods from XAI can be combined with classic class-imbalance methods to deliver better predictive performance in this climate-change domain, one of the most important application problems currently facing humanity. Here, we propose the *Counterfactual-Based SMOTE* method (CFA-SMOTE) and test it against other benchmark methods in counterfactual augmentation (i.e., DiCE-SMOTE, a SMOTE variant of DiCE) and class-imbalance augmentation (SMOTE, B-SMOTE, G-SMOTE) in our focal climate-change domain. Specifically, these methods are tested by augmenting a farming-dataset for crop-growth prediction, with synthetic data-points representing outlier, climate-events that are introduced into the dataset to improve predictive performance. In this section, we describe how CFA-SMOTE operates in this problem domain with a walkthrough (see Section 3.1) before detailing the algorithm (see Section 3.2) and explaining how it differs from other augmentation approaches (see Section 3.3).

## 3.1 Walkthrough: How CFA-SMOTE Works

Consider how CFA might operate in a SmartAg domain to generate synthetic outliers, generated as counterfactuals in an outlier-region, to populate the dataset with many extreme-weather events (see also Fig. 4). Imagine in July of 2016 on Farm-A it is warm and there is plenty of rain, good sun and, consequently, grass growth is high (call it the Farm-A$^{high}$ case), whereas in July 2018 on the same farm there is no-rain, good sun and grass growth is consequently low (as the grass has been burnt off; call it the Farm-A$^{low}$ case). In an historical dataset, these two data-points are a counterfactually-related pair, *cf(Farm-A$^{high}$, Farm-A$^{low}$)*; this known counterfactual tells us that different outcomes (low or high grass growth) occur when three features differ (i.e., rainfall, temperature and year), while other features remain constant (e.g., farm-id, county, sun, etc.). In other words, these counterfactuals, *cf(Farm-A$^{high}$, Farm-A$^{low}$)*, pair cases either side of $2\sigma$ climate boundary. Keane and Smyth [19] called these *native counterfactuals*, as they already exist in the dataset; in one sense, they can be thought of as being like a local rule, which says "when rain is low, other things being equal, grass growth will decrease". When this counterfactual method is used for data augmentation, these native counterfactuals can act as templates to create new data-points. For example, if we have an instance about another farm with high-growth, Farm-B$^{high}$, that is very similar that to the Farm-A$^{high}$ case (e.g., plenty of rain, good sun, but in a different location), then the native counterfactual about Farm-A [i.e., *cf(Farm-A$^{high}$, Farm-A$^{low}$)*] can be used as a template to suggest a new, artificial data-point about Farm-B [i.e., Farm-B$^{low}$]. Farm-B$^{low}$ is a new, extreme-weather case recording that if there is no-rainfall on that farm (using the difference-feature-value from the Farm-A$^{low}$) then grass growth will also be low. Farm-B$^{low}$ will retain the matching-features of the original Farm-A$^{high}$ case, but will have the no-rain difference-feature from the extreme-weather Farm-A$^{low}$ case (see Fig. 5 for a more abstract description). CFA-SMOTE generates these synthetic data-points for all instances in the majority class, that do not already take part in native counterfactuals. Then, these generated synthetic data-points are processed using SMOTE to further augment the minority class. This two-step augmentation technique is the key novelty in CFA-SMOTE, that is tested in the current work. Although SMOTE can re-balance the original dataset, it has some drawbacks (see Section 2.2). For example, it generates new minority instances without reference to the majority class. Also, SMOTE may introduce values that are far from the decision boundary, that may contribute little to model performance. Accordingly, CFA-SMOTE is proposed as an extension to the SMOTE algorithm that improves on its operation by combining it with a counterfactual method to improve the density of borderline area.

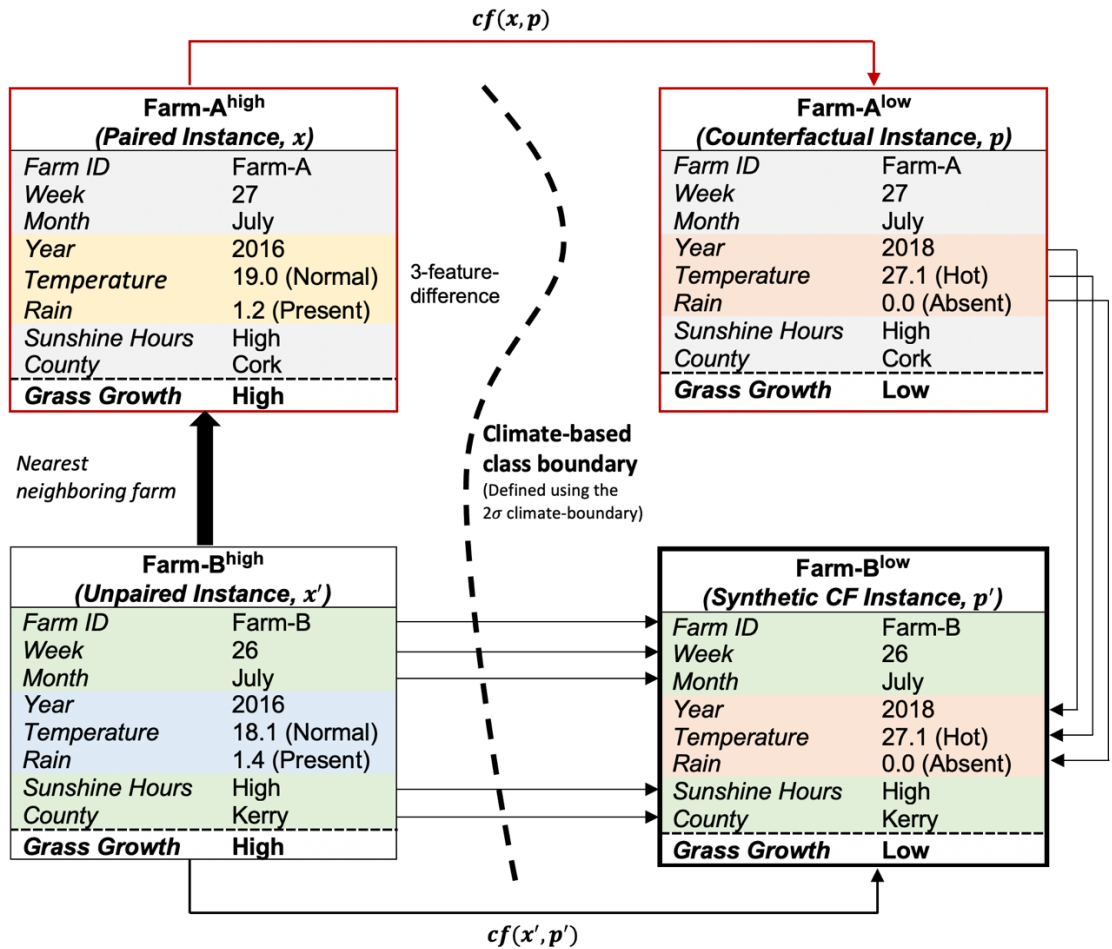

**Fig. 4** A worked example showing how CFA operates in SmartAg domain by augmenting instances for crop-growth prediction

In the present experiments, we first apply this counterfactual method to all the normal-climate data-points in the majority class (i.e., all those instances that do not already take part in known natives) to generate counterfactual extreme-weather data-points in the minority class (e.g., based on using historical natives with few feature-differences as templates). In this way, the minority class becomes populated with many new synthetic, counterfactual data-points that are "plausible" by virtue of being patterned on relationships seen in the past between historical instances (i.e., the natives). Second, we then apply the SMOTE algorithm to these new synthetic counterfactuals in the minority class. Our experiment then tests whether these new instances improve performance when future, climate-disruptive events are encountered. In the next section, we provide more detail on this new method we proposed.

## 3.2 The Algorithm: CFA-SMOTE

Counterfactual-Based SMOTE (CFA-SMOTE) is a new technique for generating synthetic instances in the minority class that combines two different strands of AI research; data-level sampling techniques for the class imbalance problem (i.e., SMOTE) and counterfactual methods for data augmentation (i.e., an instance-based counterfactual method). This method applies SMOTE only to the synthetic counterfactual instances of the minority class generated by the counterfactual technique instead of applying it to all instances of the class (as SMOTE normally does). CFA-SMOTE consists of two main stages that works as follow:

**Stage 1: Counterfactual Method Steps**

The first stage of CFA-SMOTE, applies the steps from the counterfactual algorithm, as follows:

- *Compute the native counterfactuals for the dataset (CF-Set)*. Initially, CFA-SMOTE computes all native counterfactuals $cf(x, p)$, over the whole dataset $T$. These native counterfactuals pair an instance in the majority class, $x$ (called the *paired* instance) and its counterfactually-related instance in the minority class, $p$ (called the *counterfactual* instance), where:

$$cf(x, p) \Leftrightarrow target(x_i) \neq target(p_i) \tag{1}$$

Note: They are called *native* because they already exist in the dataset $T$. Each of these native pairs has a set of *match-features* and a set of *difference-features*, where the differences determine the class change over the decision boundary.

- *Compute unpaired instances $x'$*. In this step, CFA-SMOTE filters all remaining instances in the majority class, not in native counterfactual pairs, and puts them into the *Unpaired* set:

$$x' \leftarrow (x' \in Maj_{class}) \wedge (x' \notin cf(x,p)) \quad (2)$$

It should be noted that the synthetic counterfactuals generated by the first stage of the CFA-SMOTE come from these instances in *Unpaired* set that are not already counterfactually-related to instances in the minority class (hence, they called *unpaired* instances).

- *Use the k-NN algorithm to find the nearest-neighbour paired instance $x$*. For each unpaired instance $x'$, from the majority class, CFA-SMOTE uses the *k*-NN to find its nearest-neighbour paired instance $x$, from the same class, involved in a native counterfactual pair $cf(x,p)$. Euclidean distance metric is used in calculating these nearest neighbours:

$$Distance\ (p,q) = \sqrt{\sum_{i=1}^{m}(p_i - q_i)^2} \quad (3)$$

- *Generate the synthetic counterfactual $p'$*. In the third step, having identified a candidate native counterfactual $cf(x,p)$ for $x'$, the CFA-SMOTE generates a synthetic counterfactual instance $p'$ in the minority class, using feature-values from $x'$ and $p$ and put into a set *SYN-CF*. This counterfactual generation step is done by transferring feature-values from $p$ to $p'$ and from $x'$ to $p'$, as follows:

    - For each of the *difference-features* between $x$ and $p$, take the values from $p$ into the synthetic counterfactual case $p'$.
    - For each of the *match-features* between $x$ and $p$, take the values from $x'$ into the new counterfactual case $p'$.

- *The instances in SYN-CF set are the synthetic counterfactual instances of the minority class generated by the counterfactual method*:

$$SYN\text{-}CF = \{p'_1,\ p'_2,\ p'_3, \ldots, p'_i\} \quad (4)$$

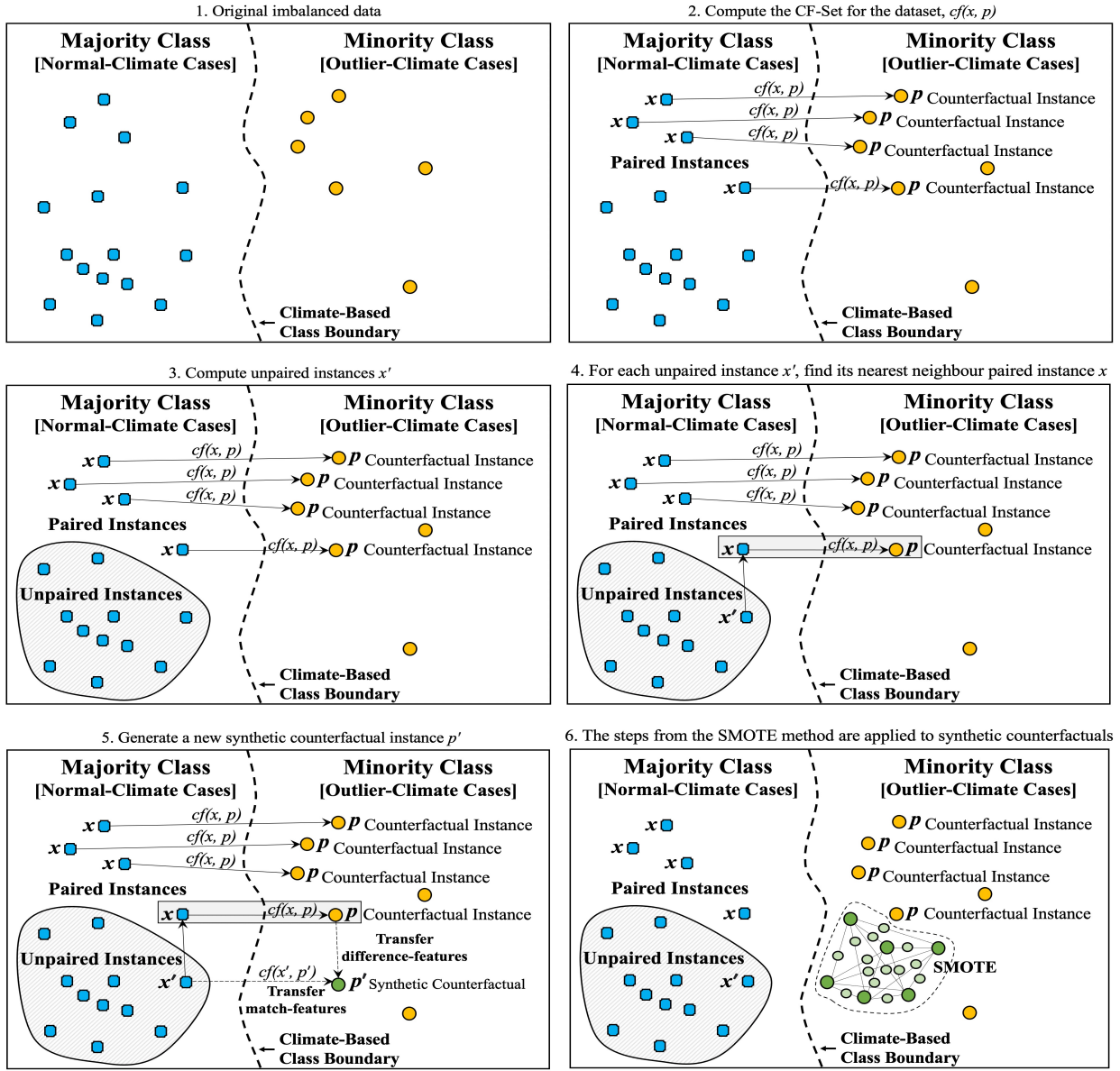

**Fig. 5** (1) An instance-based counterfactual method generates synthetic counterfactual instances (green *p'*) in the minority class (yellow circles) of outlier-climate cases. Given an instance (*x'*) in the majority class of normal-climate cases (blue crosses), it finds historical counterfactual pairs [the grey rectangles, as in *cf(x, p)*], in which *x* is a nearest neighbour of *x'*, and then uses *p* from this native-counterfactual pair as a template to generate the new, synthetic case (*p'*). (2) Then, the new synthetic counterfactuals (green *p'*) are used by SMOTE for oversampling to generate more minority instances

**Stage 2: SMOTE Method Steps**

In the second stage of the CFA-SMOTE method, the steps from the original SMOTE method are applied to each instance in *SYN-CF*, to generate the final set of the new instances in the minority class:

- *Compute the nearest-neighbour m.* For each instance $p'_i$ in *SYN-CF*, compute $m$ (where $m \in$ *SYN-CF*), as a nearest neighbour of the synthetic counterfactual instance $p'_i$.
- *Select a random nearest neighbour, $m'$* (where $m' \in m$).
- *Finally, generate a new instance $p'_{new}$.* For each $p'$ and $m'$, generate the new instances in the minority class, as follows:

$$p'_{new} \leftarrow p' + (m' - p') \times \delta, \qquad \text{where } \delta \in [0,1] \tag{5}$$

---

**ALGORITHM 1: COUNTERFACTUAL-BASED SMOTE (CFA-SMOTE)**

---

**Input:** $T$, the training set
**Output:** $p'_{new}$, a set of synthetic instances after applying CFA-SMOTE

```
1   Divide T into majority x and minority p subsets:
2       T = x ∪ p
3   Compute the CF-Set, cf(x,p) where cf(x,p) ⇔ target(x_i) ≠ target(p_i):
4       foreach x ∈ T do
5           Use 1-NN to find its nearest neighbour, p ∈ T
6       end
7   Filter unpaired instances x':
8       x' ← (x' ∈ Maj_class) ∧ (x' ∉ cf(x,p))
9   Find the nearest-neighbour paired instance, x for each an unpaired instance, x':
10      foreach x' ∉ cf(x,p) do
11          Use 1-NN to find its nearest neighbour, x ∈ cf(x,p)
12      end
13  Generate a synthetic counterfactual instance, p' for x':
14      foreach x', cf(x,p) do
15          Transfer difference-feature values from p to p'
16          Transfer match-features features values from x' to p'
17      end
18  Apply SMOTE to the counterfactual instances, p'
19      foreach p' ∈ T do
20          Compute m ∈ p', as a nearest neighbour of the counterfactual instance p' in T
21          Choose a random sample, m' ∈ m
22          Create a new instance, p'_new
23              foreach p', m' do
24                  p'_new ← p' + (m' - p') × δ, δ ∈ [0,1]
25              end
26      end
27  return p'_new
```

Note, that these new synthetic data points $p'_{new}$ are generated along the line between the minority counterfactual instances and their nearest neighbors of the same class, thus improving the density of borderline area. In the next section, we test CFA-SMOTE on a farming-dataset for crop-growth prediction, that creates synthetic data-points representing outlier, climate-events that are introduced to the dataset with a view to improving predictive performance.

### 3.3 How CFA-SMOTE Differs from Other Data Augmentation Techniques?

This hybrid CFA-SMOTE method is quite different from SMOTE-variants in several significant respects. First, it does not use the known instances in the minority class for oversampling (e.g., as in SMOTE) but rather only uses the synthetic counterfactuals generated by the counterfactual method for generating the new minority instances. In other words, the synthetic counterfactuals are the sole inputs for SMOTE algorithm. Second, CFA-SMOTE does not rely on the topology of the majority class (e.g., as in SWIM [1]), but acts in a very local way using the counterfactual relation between a single majority instance and a minority one. As such, this method is also different from other popular constraint-optimization counterfactual techniques (e.g., DiCE) which filters results based on broad constraints of proximity and diversity (rather than operating more locally), sometimes producing out-of-distribution, invalid data-points. Note, we include a SMOTE-variant of DiCE in the current tests to determine whether it is the specific properties of our method that leads to the results found. Third, Temraz et al.'s [37] used an instance-based counterfactual method directly to augment the climate-change dataset, finding weak-support for overall performance improvements; it did not include the SMOTE-steps used in the current algorithm.

## 4 Experiment 1: Testing CFA-SMOTE on Climate Change

Grass-growth predictions from datasets modified by several augmentation methods -- SMOTE, Borderline-SMOTE, Geometric-SMOTE, DiCE-SMOTE, CFA-SMOTE -- were compared against a Baseline control (with no data augmentation) on the climate-disrupted year of 2018; specifically, concentrating on the key climate-disrupted months of March, July and October in that year (see Fig. 1). Each method was run on 12 different datasets, with systematically varied Imbalance Ratios (IR; N-of- Majority-Instances / N-of-Minority-Instances) that were either Extreme / Moderate /

Mild (see Table 1). At present, most historical datasets are probably moderate-to-severe in terms of their class-imbalance ratios (IR=5-100) as they probably encode few extreme-weather instances. This critical manipulation has not been done before, as a test to determine the most effective augmentation method, the one that works well with high-imbalance datasets. Predictions were measured using the Mean Absolute Error in grass-growth predictions for the key months in 2018.

## 4.1 Setup & Methods: Techniques

Six conditions were use in the study covering a no-augmentation baseline (Baseline), and augmentations from the five target methods: SMOTE [4, 9], Borderline-SMOTE [10], Geometric-SMOTE [8], DiCE-SMOTE (a modification of Mothilal et al., [27]) and our novel CFA-SMOTE (a modification of Temraz et al., [37]). Specifically, the tested techniques were as follows:

- **Synthetic Minority Over-sampling TEchnique (SMOTE) [4]:** SMOTE is one of the most widely used techniques to balance the class distribution of datasets. In the present experiment, we used the available SMOTE implementation provided by the *imbalanced-learn* Python library to generate synthetic instances in the minority class, to balance each dataset.
- **Borderline-SMOTE (B-SMOTE) [10]:** is another widely used solution to the class-imbalance problem. This method only oversamples the minority class instances (called *borderline*) near the decision boundary. In the present experiment, we generated Borderline-SMOTE data using the standard existing *scikit-learn* python implementation of Borderline-SMOTE.
- **Geometric-SMOTE (G-SMOTE) [8]:** is a more recent solution to the class-imbalance problem. It defines a geometric region by identifying a safe area around each minority class instance and generates synthetic instances in this area. In a similar vein, we used the Python implementation of Geometric SMOTE algorithm which is compatible with *scikit-learn* and *imbalanced-learn*, to balance each dataset.
- **DiCE-SMOTE (a modification of Mothilal et al., [27]):** DiCE is a benchmark counterfactual technique using perturbation to generate a set of diverse counterfactual instances. To make the comparison fair from a counterfactual perspective, we also combined a different XAI-counterfactual method, DiCE, with the SMOTE algorithm; specifically, for each instance in the majority class, we first run DiCE to generate a counterfactual instance in the minority class. Then SMOTE algorithm was applied on these synthetic counterfactuals generated by DiCE, to mimic the steps taken in our CFA method. For the DiCE implementation, we used an open-source Python library named *dice-ml*, designed by the Microsoft research team, a library is supported by *scikit-learn*.

- **Counterfactual-Based SMOTE (CFA-SMOTE):** Finally, we applied the novel CFA-SMOTE. As we did with DiCE-SMOTE, our instance-based counterfactual method was first applied to a dataset to generate a set of synthetic counterfactual instances in the minority class, and then these synthetic counterfactual instances are input to SMOTE to generate further instances in the minority class.

Finally, it should be apparent that the main difference between CFA-SMOTE, DiCE-SMOTE and other SMOTE-variants is that in CFA-SMOTE and DiCE-SMOTE, use a counterfactual method to supplement SMOTE-augmentations. Clearly, if we find that CFA-SMOTE and DiCE-SMOTE differ in their performance, then we know that it is the instance- or perturbation-based aspects of these methods makes a difference.

**Table 1** PBI dataset versions were created different Imbalance Ratios (IR), broadly divided into Mild, Moderate and Extreme; Moderate-Extreme IRs may better represent current climate-related datasets

| DataSets | Majority (Normal-Cases) | Minority (Outlier-Cases) | Imbalance Ratio | Degree of Imbalance |
|---|---|---|---|---|
| D1 | 6,000 | 3,810 | 1.5 | Mild |
| D2 | 8,000 | 3,810 | 2.0 | Mild |
| D3 | 13,000 | 3,810 | 3.4 | Mild |
| D4 | 15,000 | 3,810 | 3.9 | Mild |
| D5 | 32,719 | 3,810 | 8.5 | Moderate |
| D6 | 32,719 | 3,000 | 10.9 | Moderate |
| D7 | 32,719 | 1,800 | 18.1 | Moderate |
| D8 | 32,719 | 1,000 | 32.7 | Moderate |
| D9 | 32,719 | 600 | 54.5 | Extreme |
| D10 | 32,719 | 400 | 81.7 | Extreme |
| D11 | 32,719 | 275 | 118.9 | Extreme |
| D12 | 32,719 | 200 | 163.5 | Extreme |

## 4.2  Setup & Methods: Datasets

In the present study, each method was applied to 12 different datasets, that were systematically varied in their Imbalance Ratio (IR) as Extreme (163.5, 118.9, 81.7, 54.5) or Moderate (32.7, 18.1, 10.9, 8.58) or Mild (3.93, 3.41, 2,09, 1.57) (see Table 1).  The 12 datasets were created by randomly sampling from the original PBI dataset to change the majority-class size (from 32,719-6,000 instances) and minority classes (from 3810-200 instances).   The original PBI dataset was drawn from 6,000+ farms over 6 years (i.e., 2013-2018) giving a dataset of N=70,091 instances, after data cleaning. The training data for the test (based on 2013-2016) has N = 36,529 cases of which 89.6% (N=32,719) were normal-weather, majority-class instances and 10.4% (N=3,810) were extreme-weather, minority-class instances (giving an IR= 8.58) defined using the $2\sigma$ climate-boundary. The test-data used the years 2017 and 2018 (though, 2018 is only reported here, similar but weaker effects were found for 2017 as it had less climate disruption than 2018).   Prediction accuracy was measured using Mean Absolute Error (MAE) were *AE=|actual-grass-growth - predicted-grass-growth|*. The standardized grass-growth measure is kilograms of dry matter grown per hectare, per day (kg/DM/ha/day). In the tests, we oversample the minority class using each of the augmentation methods until we have the same number of instances in each class creating fully-balanced datasets.  A *k*-NN model was used to predict grass growth over the augmented datasets (similar results were found using Linear Regression and a Multilayer Perceptron, that are not reported here).

## 4.3  Results & Discussion

The results of the experiment generally show that data augmentation helps to improve accuracy (reducing MAE) in the climate-disrupted months of 2018 (i.e., March, July, October) and that the CFA-SMOTE works best (see Fig. 6 and Tables 2-4 for details).  Also, the benefits seen for data augmentation methods relative to the no-augmentation baseline are most pronounced in datasets that have moderate to extreme imbalance ratios. Note, that overall, data augmentation only improves accuracy for the climate-disrupted months, it does not impact the other "normal" months (this data is not reported).  Fig. 6 graphs the MAE results for the five data augmentation methods (i.e., SMOTE, B-SMOTE, G-SMOTE, DiCE-SMOTE and CFA-SMOTE) against the Baseline (with no augmentation to the dataset). CFA-SMOTE outperforms the Baseline, SMOTE, G-SMOTE, B-SMOTE and DiCE-SMOTE with lower MAE values when the degree of imbalance is extreme and moderate. For example, the MAE for March 2018 shows that CFA-SMOTE (MAE=24.9 kg/DM/ha) performed better than Baseline (MAE=38.6kg/DM/ha), SMOTE (MAE=35.2kg/DM/ha), B-SMOTE (MAE=34.5kg/DM/ha), G-SMOTE (MAE=29.3kg/DM/ha) and DiCE-SMOTE (MAE=43.0kg/DM/ha) when the degree of imbalance is extreme (IR=163.5). Similarly, when the degree of imbalance is moderate (IR=32.7), the MAE for March 2018 shows that CFA-SMOTE (MAE=23.9 kg/DM/ha) performed better than Baseline (MAE=39.9kg/DM/ha), SMOTE (MAE=35.8kg/DM/ha), B-SMOTE (MAE=35.8kg/DM/ha), G-SMOTE (MAE=31.7kg/DM/ha) and DiCE-SMOTE (MAE=41.1kg/DM/ha).

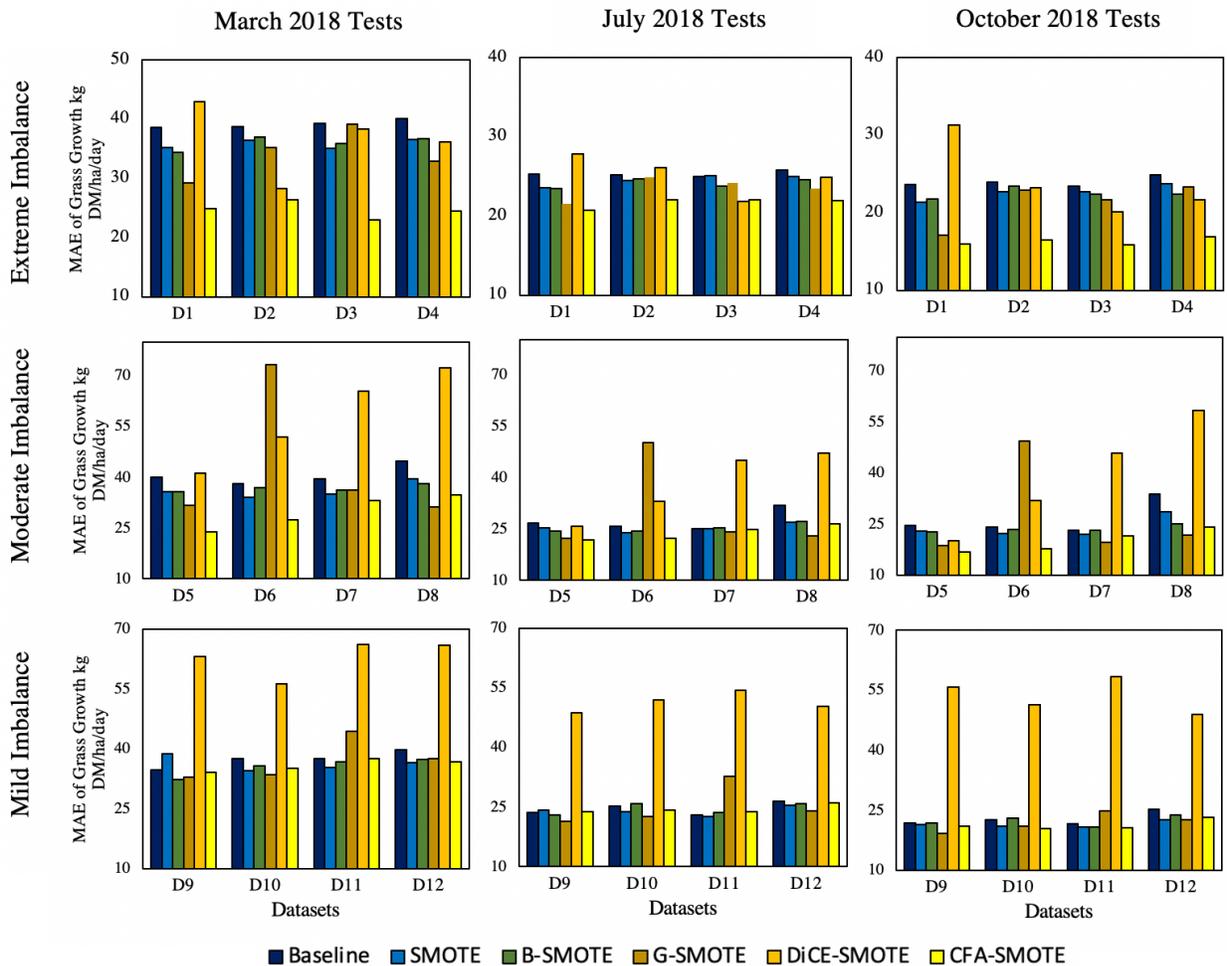

**Fig. 6** Mean Absolute Error (MAE) in predicted grass growth for March, July, and October of 2018, the climate-disrupted months, using datasets with differing Imbalance Ratios (IRs), grouped as Mild, Moderate or Extreme, for five data augmentation methods -- SMOTE, B-SMOTE, G-SMOTE, DiCE-SMOTE and CFA-SMOTE - compared to the Baseline (no augmentation)

Overall, for March 2018 tests, the CFA-SMOTE does best in 7/12 datasets, with the SMOTE and G-SMOTE doing best in 2 datasets and B-SMOTE just 1. For July 2018 tests, CFA-SMOTE and G-SMOTE do better than all the other methods in 5 out of 12 datasets for each method, with SMOTE and DiCE-SMOTE being the next best with one dataset for each. For October 2018 tests, CFA-SMOTE doing better in 8 out of 12 datasets, with G-SMOTE being the next best with 4 datasets.

**Table 2** MAE for March 2018 for each method by degree of imbalance (i.e., Extreme, Moderate, Mild)

| Method | Degree of imbalance: Extreme | | | | Degree of imbalance: Moderate | | | | Degree of imbalance: Mild | | | |
|---|---|---|---|---|---|---|---|---|---|---|---|---|
| | IR= 163.5 | IR= 118.9 | IR= 81.7 | IR= 54.5 | IR= 32.7 | IR= 18.1 | IR= 10.9 | IR= 8.5 | IR= 3.9 | IR= 3.4 | IR= 2.0 | IR= 1.5 |
| Baseline | 38.6 | 38.8 | 39.3 | 40.1 | 39.9 | 38.1 | 39.5 | 44.7 | 34.8 | 37.6 | 37.6 | 39.7 |
| SMOTE | 35.2 | 36.5 | 35.1 | 36.6 | 35.8 | 34.0 | 35.1 | 39.5 | 38.8 | 34.6 | **35.4** | **36.6** |
| B-SMOTE | 34.5 | 37.0 | 36.0 | 36.7 | 35.8 | 37.0 | 36.3 | 38.1 | **32.3** | 35.7 | 36.8 | 37.4 |
| G-SMOTE | 29.3 | 35.3 | 39.2 | 33.0 | 31.7 | 73.2 | 36.3 | **31.3** | 33.0 | **33.6** | 44.4 | 37.6 |
| DiCE-SMOTE | 43.0 | 28.3 | 38.4 | 36.2 | 41.1 | 51.9 | 65.3 | 72.4 | 63.2 | 56.2 | 66.2 | 65.9 |
| CFA-SMOTE | **24.9** | **26.4** | **23.1** | **24.5** | **23.9** | **27.4** | **33.1** | 34.8 | 34.1 | 35.1 | 37.6 | 36.8 |

**Table 3** MAEs for July 2018 for each method by degree of imbalance (i.e., Extreme, Moderate, Mild)

| Method | Degree of imbalance: Extreme | | | | Degree of imbalance: Moderate | | | | Degree of imbalance: Mild | | | |
|---|---|---|---|---|---|---|---|---|---|---|---|---|
| | IR= 163.5 | IR= 118.9 | IR= 81.7 | IR= 54.5 | IR= 32.7 | IR= 18.1 | IR= 10.9 | IR= 8.5 | IR= 3.9 | IR= 3.4 | IR= 2.0 | IR= 1.5 |
| Baseline | 25.3 | 25.2 | 25.0 | 25.8 | 26.7 | 25.7 | 25.0 | 31.8 | 23.6 | 25.3 | 22.9 | 26.5 |
| SMOTE | 23.6 | 24.5 | 25.1 | 25.0 | 25.3 | 23.8 | 25.1 | 26.9 | 24.3 | 23.9 | **22.5** | 25.4 |
| B-SMOTE | 23.5 | 24.7 | 23.8 | 24.6 | 24.4 | 24.2 | 25.3 | 27.1 | 22.9 | 25.8 | 23.7 | 25.8 |
| G-SMOTE | 21.6 | 24.9 | 24.2 | 23.5 | 22.1 | 50.1 | **24.1** | 23.0 | 21.4 | 22.5 | 32.7 | **24.1** |
| DiCE-SMOTE | 27.9 | 26.1 | **21.9** | 24.9 | 25.7 | 32.9 | 44.9 | 47.1 | 48.7 | 51.9 | 54.4 | 50.3 |
| CFA-SMOTE | **20.8** | **22.1** | 22.1 | **22.0** | **21.7** | **22.1** | 24.8 | 26.5 | 23.8 | 24.3 | 23.9 | 26.0 |

**Table 4** MAE for October 2018 for each method by degree of imbalance (i.e., Extreme, Moderate, Mild)

| Method | Degree of imbalance: Extreme | | | | Degree of imbalance: Moderate | | | | Degree of imbalance: Mild | | | |
|---|---|---|---|---|---|---|---|---|---|---|---|---|
| | IR= 163.5 | IR= 118.9 | IR= 81.7 | IR= 54.5 | IR= 32.7 | IR= 18.1 | IR= 10.9 | IR= 8.5 | IR= 3.9 | IR= 3.4 | IR= 2.0 | IR= 1.5 |
| Baseline | 23.7 | 24.0 | 23.5 | 24.9 | 24.5 | 24.1 | 23.2 | 33.8 | 21.8 | 22.6 | 21.5 | 25.1 |
| SMOTE | 21.4 | 22.7 | 22.7 | 23.8 | 22.9 | 22.2 | 22.1 | 28.7 | 21.3 | 20.9 | 20.8 | 22.6 |
| B-SMOTE | 21.8 | 23.5 | 22.4 | 22.4 | 22.6 | 23.4 | 23.2 | 25.0 | 21.8 | 23.0 | 20.7 | 23.7 |
| G-SMOTE | 17.2 | 23.0 | 21.7 | 23.4 | 18.6 | 49.4 | **19.6** | **21.8** | **19.1** | 21.0 | 24.7 | **22.5** |
| DiCE-SMOTE | 31.3 | 23.3 | 20.2 | 21.7 | 20.2 | 31.8 | 45.9 | 58.5 | 55.8 | 51.4 | 58.4 | 49.0 |
| CFA-SMOTE | **16.0** | **16.5** | **15.9** | **16.9** | **16.8** | **17.8** | 21.5 | 24.1 | 21.0 | **20.4** | **20.6** | 23.1 |

Fig. 7 shows the average MAE scores across all datasets obtained by each method in the climate-disrupted months of 2018 (i.e., March, July, October). CFA-SMOTE achieved greater improvements in the results by comparison to the other methods and the baseline (i.e., SMOTE, B-SMOTE, G-SMOTE, DiCE-SMOTE and Baseline). For example, the average MAE across the October test-set showed that CFA-SMOTE (avg. MAE=19.2 kg/DM/ha) performed better than DiCE-SMOTE (avg. MAE=38.9 kg/DM/ha), Baseline (avg. MAE=24.3 kg/DM/ha), G-SMOTE (avg. MAE=23.5 kg/DM/ha), B-SMOTE (avg. MAE=22.7 kg/DM/ha), and the SMOTE (avg. MAE=22.6 kg/DM/ha) conditions. These improvements in the results are tested using the non-parametric Wilcoxon signed-rank test [39] to determine whether the methods were significantly different. For each comparison we compute the *p*-value for March, July and October tests. Table 5 shows the results of applying Wilcoxon's test in pairwise comparisons between CFA-SMOTE and SMOTE, B-SMOTE, G-SMOTE, DiCE-SMOTE and Baseline. The results show that CFA-SMOTE produced statistically significant improvements in the MAEs in all but three of comparisons. These results show that the counterfactual method can generate "plausible" artificial data-points that improve prediction for test items in a minority class, as it re-uses counterfactual-relationships for historical data, using past counterfactuals as templates for new synthetic data.

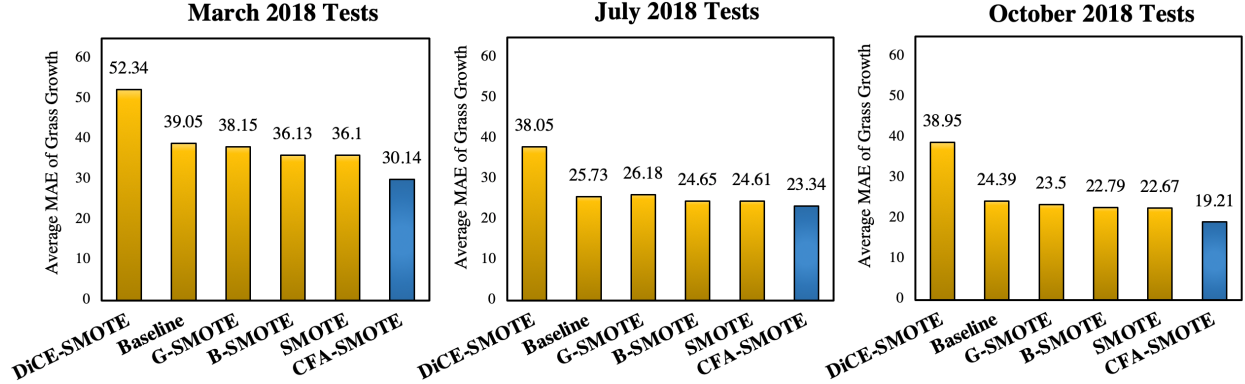

**Fig. 7** Average Mean Absolute Errors (avg. MAE) of grass growth (measured in kg DM/ha/day) across all datasets for the climate-disrupted months of March, July, and October in 2018

**Table 5** Wilcoxon's signed rank test results for the comparison of CFA-SMOTE against other methods from MAEs scores obtained by each test (i.e., March, July and October)

| Comparison CFA-SMOTE vs. | March 2018 Tests | | July 2018 Tests | | October 2018 Tests | |
|---|---|---|---|---|---|---|
| | *p*-value | Significance | *p*-value | Significance | *p*-value | Significance |
| Baseline | 0.003 | True** | 0.006 | True** | 0.000 | True*** |
| SMOTE | 0.009 | True** | 0.052 | False$^{NS}$ | 0.002 | True** |
| B-SMOTE | 0.009 | True** | 0.012 | True* | 0.000 | True*** |
| G-SMOTE | 0.020 | True* | 0.569 | False$^{NS}$ | 0.129 | False$^{NS}$ |
| DiCE-SMOTE | 0.000 | True*** | 0.000 | True*** | 0.000 | True*** |

Level of significance (α):   *=0.05    **=0.01    ***=0.001    NS=Not significant

## 4.4 Computational Complexity Analysis

For completeness, in this section we consider the computational complexity of the successful CFA-SMOTE method by analyzing each one of its steps. First, assume that $T$ is the training data and $n$ the total number of instances. As described in Section 3, the first step of the CFA-SMOTE is to compute the native counterfactuals in $T$ using the *k*-NN algorithm (Lines 3-6 of Algorithm 1). Native counterfactuals reflect a counterfactual relation between existing instances in $T$, that are in opposing classes either side of a decision boundary (i.e., for each instance in the majority class, it finds its nearest neighbour from the minority class). So, the time complexity for this step is of $O(n*k)$; where $n$ is the number of instances in the majority class and $k$ is the nearest neighbour of that instance. Step 2 computes/filters a set *Unpaired* (i.e., those are remaining majority class instances that are not taking part in the native counterfactuals) (Lines 7-8 of Algorithm 1); so, the time complexity for this step is $O(n)$ because all majority class instances have to be examined to determine whether are taking part in the native counterfactuals or not – if there are twice as many instances, it takes twice as long. In Step 3 (Lines 9-12 of Algorithm 1), for each instance in *Unpaired* set, CFA-SMOTE uses a *k*-NN to find its nearest-neighbour paired instance in the majority class. This step has a complexity of $O(n*k)$, where $n$ is the number of instances and $k$ is its nearest neighbour on the same class. Step 4 applies CFA to generate/computes new counterfactual instances in the minority class using values from the *match* and *difference* features (Lines 13-17 of Algorithm 1). This step has a complexity of $O(match+diff)$. In step 5 (Lines 18-26 of Algorithm 1), the SMOTE algorithm is applied on the synthetic counterfactuals generated by CFA. In SMOTE, for each instance in the minority class, it finds its *k*-nearest minority class neighbours. So the time complexity is $O(n*k)$. Then, it generates a new synthetic instance in the minority class between each minority class seed and its random neighbour in constant time. So, the time complexity is $O(1)$ which means it is constant.

The original SMOTE is known to have a low runtime complexity (≈0.46 second) (see Fig. 8). In a comparison with SMOTE, the complexity of CFA-SMOTE (≈8.85 second) is higher due to its hybrid approach (since it applies two different techniques: CFA and SMOTE). However, the difference in the runtime between both methods is set to 8.3853 seconds, which is still acceptable, due to the fact that CFA-SMOTE takes into consideration some additional mechanisms to better overcome the imbalance problem (e.g., finding native counterfactuals, filtering unpaired instances, and computing the synthetic counterfactuals) making it at the same time a bit more complex. In contrast, the advantage of CFA-SMOTE over other optimization counterfactual methods (e.g., DiCE and DiCE-SMOTE) is that the computational complexity of the CFA-SMOTE is low. Because these optimization methods introduce various complex operations to preprocess the training data, resulting in the increased computational overheads. For example, DiCE perturbs feature-values and filters results based on broad constraints of proximity and diversity (see Fig. 8).

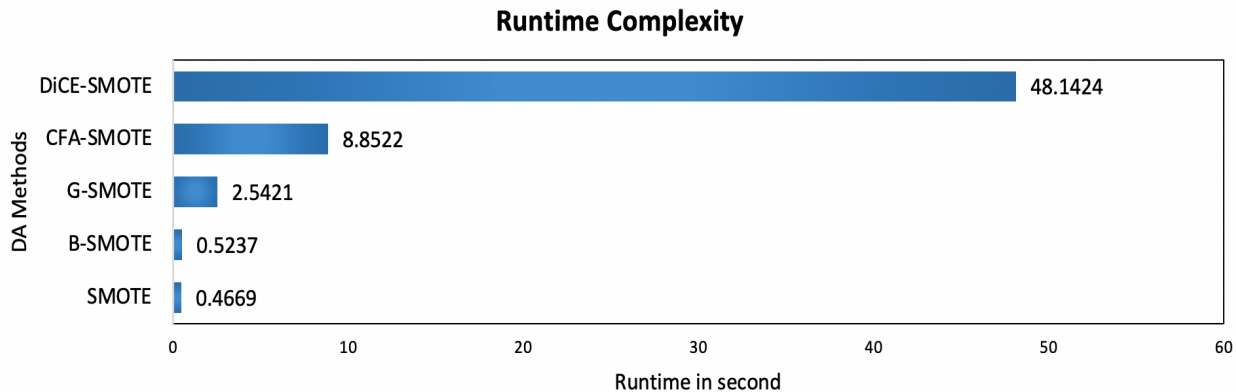

**Fig. 8** Average runtime complexity of each data augmentation method across datasets

## 5 Conclusions

Climate change is currently presenting humanity with one of its most severe challenges. Worryingly, the apparent utility of advanced AI technologies to these problems, is undermined by the out-of-distributional aspects of climate change itself. Two significant novelties are introduced in this paper. The first novelty hinges on the claim that the predictive problems that AI faces on climate change can be cast as class-imbalance issues; that is, prediction from historical datasets is difficult because, by definition, they lack sufficient minority-class instances of "climate outlier events". The second novelty involved the introduction of a new data augmentation method -- called Counterfactual-Based SMOTE (CFA-SMOTE) – that combines an instance-based counterfactual method (CFA) with the well-known SMOTE class-imbalance method. We report an extensive experiment using this CFA-SMOTE method, comparing it to benchmark counterfactual and class-imbalance methods under different conditions (i.e., class-imbalance ratios). The present novel results show that there is a way out of the i.i.d. assumption that afflicts AI methods dealing with climate-change phenomena, that data augmentation techniques can help to overcome the distributional inadequacies facing these techniques. Furthermore, it appears that our proposed novel counterfactual method works well, especially when imbalance ratios are higher. So, we would argue for a cautious optimism in approaching these challenges, though it is still clear that a considerable amount of work needs to be done.

## Acknowledgments

This publication has emanated from research conducted with the financial support of (i) Science Foundation Ireland (SFI) to the Insight Centre for Data Analytics under Grant Number: 12/RC/2289_P2 and (ii) SFI and the Department of Agriculture, Food and Marine on behalf of the Government of Ireland to the VistaMilk SFI Research Centre under Grant Number: 16/RC/3835.## Data availability

The datasets generated during and/or analysed during the current study are available from the corresponding author on reasonable request.

## Declarations

**Conflict of Interests** The authors declare that they have no known competing financial interests or personal relationships that could have appeared to influence the work reported in this paper.